\pdfoutput=1

\documentclass[11pt]{article}

\usepackage[preprint]{coling}

\usepackage{times}
\usepackage{latexsym}

\usepackage{multirow}
\usepackage{natbib}

\usepackage{graphicx}
\usepackage{booktabs}
\usepackage{amsmath}
\usepackage{float}
\usepackage{pifont}
\usepackage{amssymb}
\usepackage{bbding}
\usepackage{mdframed}
\usepackage{xcolor}
\usepackage{algorithm}
\usepackage{algorithmic}

\usepackage[T1]{fontenc}

\usepackage[utf8]{inputenc}

\usepackage{microtype}

\usepackage{inconsolata}

\usepackage{graphicx}

\title{PMSS: Pretrained Matrices Skeleton Selection for LLM Fine-tuning}

\author{
 \textbf{Qibin Wang\textsuperscript{1,4}},
 \textbf{Xiaolin Hu\textsuperscript{2,4}},
 \textbf{Weikai Xu\textsuperscript{3,4}},
 \textbf{Wei Liu\textsuperscript{4}},
 \textbf{Jian Luan\textsuperscript{4}},
 \textbf{Bin Wang\textsuperscript{4}},
\\
 \textsuperscript{1}Peking University,
 \textsuperscript{2}Gaoling School of Artificial Intelligence, Renmin University of China,
 \\
 \textsuperscript{3}University of Electronic Science and Technology of China,
 \textsuperscript{4}XiaoMi AI Lab,
\\
 \small{
   \textbf{Correspondence:} 
   {wangqibin@stu.pku.edu.cn},
   {wangbin11@xiaomi.com}
 }
}
\begin{document}
\maketitle
\begin{abstract}
Low-rank adaptation (LoRA) and its variants have recently gained much interest due to their ability to avoid excessive inference costs.
However, LoRA still encounters the following challenges:
(1) Limitation of low-rank assumption; and (2) Its initialization method may be suboptimal.
To this end, we propose PMSS(Pre-trained Matrices Skeleton Selection), which enables high-rank updates with low costs while leveraging semantic and linguistic information inherent in pre-trained weight.
It achieves this by selecting skeletons from the pre-trained weight matrix and only learning a small matrix instead.
Experiments demonstrate that PMSS outperforms LoRA and other fine-tuning methods across tasks with much less trainable parameters.
We demonstrate its effectiveness, especially in handling complex tasks such as DROP benchmark(+3.4\%/+5.9\% on LLaMA2-7B/13B) and math reasoning(+12.89\%/+5.61\%/+3.11\% on LLaMA2-7B, Mistral-7B and Gemma-7B of GSM8K).
The code and model will be released soon.
\end{abstract}


\section{Introduction}

Large language models (LLMs) have demonstrated exceptional capabilities across a wide range of natural language processing (NLP) tasks \citep{radford2019language}. The pre-training stage provides LLMs with foundational abilities for general tasks, but fine-tuning is typically required to better adapt them to specific downstream tasks \citep{dai2015semi}.
Full fine-tuning, though effective in unlocking the potential of LLMs, is resource-intensive, introducing storage and computation challenges. 
As the scale of model training data and parameters continues to grow, the expense of full fine-tuning has become increasingly prohibitive, 
hindering the adoption of LLMs in scenarios where resources are limited.

To address this issue, Parameter-Efficient Fine-Tuning (PEFT) methods have been proposed to reduce the time and computation cost for fine-tuning pre-trained models\citep{houlsby2019parameter,hu2021lora,lester2021power,li2021prefix}. 
Among these methods, LoRA\citep{hu2021lora} has gained particular success for its effectiveness and simplicity without altering model architecture or introducing any inference latency. 
However, LoRA still faces two fundamental limitations: first, its low-rank assumption may not generalize well to complex tasks, and second, its initialization method can result in slower or suboptimal convergence.

Recent studies have found that LoRA's efficacy diminishes empirically in some complex tasks, especially those that differ from the pre-training dataset compared with full fine-tuning\citep{biderman2024lora}. This phenomenon is hypothesized to stem from the inherent low-rank assumption underlying LoRA, which posits that an update of weight during fine-tuning occurs within a low-rank subspace and can be well-approximated by a low-rank matrix production. 
\citet{chen2024quanta} and \citet{jiang2024mora} have evaluated the generalizability of the low-rank assumption, showing that specific complex tasks typically exhibit a higher intrinsic rank.
Other researchers have focused on LoRA's initialization method, where adapter matrix $B$ is initialized with zeros and matrix $A$ with Gaussian noise.
PiSSA\citep{meng2024pissa} and its follow-up works\citep{balazy2024lora, lingam2024svft,wang2024milora, yang2024corda}, usually have employed low-rank approximations of the original pretrained matrices, such as low-rank SVD approximation, to initialize adapter matrices in LoRA. 
These studies demonstrate that alternative initialization methods can improve performance across different models and datasets.
The success of these methods highlights the suboptimality in LoRA, further indicating that pretrained matrices contain rich semantic content highly pertinent to various downstream tasks. 
However, these works have not further examined the pre-trained matrices themselves.

To address these challenges, we consider the two key factors simultaneously: 
(1) Overcome the limitations of low-rank assumption. Even in resource-constrained environments, it is essential to enable high-rank updates during fine-tuning to gain an advantage in handling more complex tasks, such as mathematical reasoning.
(2) Leverage the semantic and linguistic information inherent in pretrained matrices rather than initialization in Gaussian noise or zeros, bridging the gap between the pre-training and fine-tuning stages.

\begin{figure}[t]
    \includegraphics[width=\columnwidth]{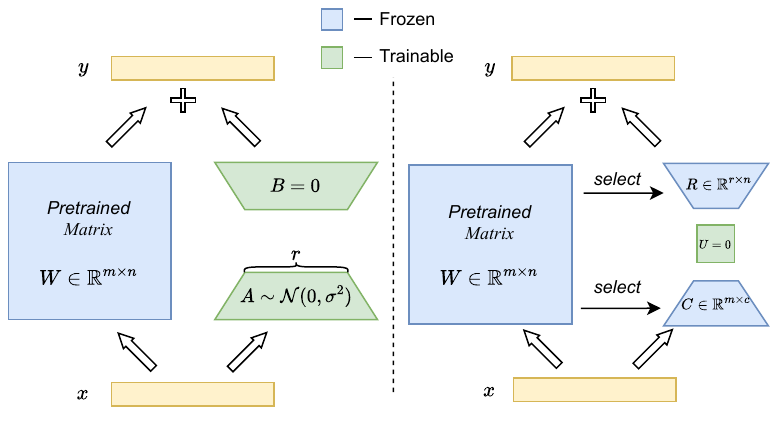}
    \caption{
    An overview of LoRA and our proposed PMSS. 
    The distinction lies in that PMSS freezes $C$ and $R$ and only updates $U$ during the fine-tuning stage.
    Note that \textit{select} denotes we select the row and column skeletons from the original pre-trained matrices to construct matrices $C$ and $R$, which ensures update happens in the subspace spanned by skeletons of the original weight. 
    Further, $C$ and $R$ can be compactly represented by one-dimensional index vectors.
    }
    \label{fig:overview}
\end{figure}

To this end, we propose PMSS(\textbf{P}re-trained \textbf{M}atrices \textbf{S}keleton \textbf{S}election), a novel parameter efficient fine-tuning method designed to enhance the parameter efficiency of large language model while leveraging the intrinsic semantic structure of pre-trained matrices. As illustrated in Figure \ref{fig:overview}, by carefully selecting the row and column skeletons and freezing them during training, we ensure that the updates occur within the subspace spanned by these components. As hypothesized by ReFT\citep{wu2024reft}, pre-training is likely the crucial stage in endowing models with capabilities, while instruction tuning acts merely as a form of style transfer. Our experiments demonstrate empirically that after pre-training on extensive and diverse datasets, over-parameterized models have already been positioned into a subspace that captures a wide range of linguistic and semantic patterns, meaning that only minimal adjustments are needed to adapt the model to specific downstream tasks.

The summary of our contributions is as follows:

\begin{itemize}
    \item We introduce a novel fine-tuning method while preserving the intrinsic semantic structure of pre-trained matrices and enabling high-rank updates. Our method further reduces the number of trainable parameters compared to the state-of-the-art LoRA. 
    
    \item We compare our method with LoRA and other parameter-efficient adaptation methods on the DROP, commonsense reasoning, and math reasoning benchmarks. Our method yields better results compared to LoRA, especially on complex tasks such as DROP(+3.4\%/+5.9\% on LLaMA2-7B/13B) and math reasoning(+12.89\%/+5.61\%/+3.11\% on LLaMA2-7B, Mistral-7B and Gemma-7B of GSM8K). 
    
    \item Through our experiments, we demonstrate that fine-tuning happens in tiny subspaces related to subspaces spanned by skeletons of model parameters.
\end{itemize}

\section{Related Work}

\subsection{Intrinsic Dimension and Subspace Learning}

\citet{li2018measuring} first introduced the intrinsic dimension of objective landscape and demonstrated different tasks exhibit varying intrinsic dimensions through random subspace training. This work found that many tasks inherently have lower intrinsic dimensions. Following this, \citet{aghajanyan2021intrinsic} further elucidated that common pre-trained language models typically exhibit low intrinsic dimensions, with larger models often possessing even lower intrinsic dimensions. \citet{gur2018gradient} showed empirically that after a short period of training, the gradient dynamically converges to a tiny subspace, which is preserved over even long periods of training.

Based on intrinsic dimension and subspace learning, a series of methods \citep{gressmann2020improving, li2022low, li2022subspace, gauch2022few,zhang-etal-2023-tinysubspace} are proposed. Most of these works rely on either random projections or sampling from optimization trajectories to extract subspaces, allowing large-scale model training within a tiny subspace.

\subsection{Column Subset Selection, CUR and Interpolative Decomposition}
Column Subset Selection Problem(CSSP) has been extensively studied within the theoretical computer science community\citep{boutsidis2009improved,deshpande2010efficient,tropp2009column,altschuler2016greedy}, revolves around selecting a small subset of representative column skeletons of a matrix. The goal of CSSP is to identify column skeletons covering column space and capturing the essential information of the original matrix. Compared with SVD, CSSP provides a more interpretable way while preserving the underlying structure, such as sparsity and non-negativity.

Inspired by core idea of CSSP, one way to achieve low-rank matrix approximation leverages the self-expression of data, which is the notion that data is better represented by other data points rather than an abstract set of bases\citep{hamm2020perspectives}. There are two representative methods: CUR Decomposition(also named Skeleton Decomposition)\citep{mahoney2009cur} and Interpolative Decomposition(ID)\citep{cheng2005compression}. We will formally define them in Section~\ref{sec:preliminary}.

\subsection{Parameter-Efficient Fine-Tuning}

Parameter-Efficient Fine-Tuning (PEFT) methods typically only train a small fraction of parameters while keeping the vast majority of parameters frozen to adapt large-scale models to downstream tasks. LoRA\citep{hu2021lora} has merged as a prominent fine-tuning technique of large pre-trained models, offering a computation and memory-efficient alternative to full fine-tuning.  PiSSA\citep{meng2024pissa} initializes the adapter matrices using a low-rank SVD of the original weights and only updates principal singular components. LoRA-XS\citep{balazy2024lora} performs a basis adaptation for frozen principal singular values and vectors. Concurrent with our work, CURLoRA\citep{fawi2024curlora} adopted a CUR-modified method but involves the random selection of row and column skeletons with smaller norms to mitigate catastrophic forgetting. We will elucidate a comparison in detail in Section~\ref{sec:comparison}.

\section{Preliminary}
\label{sec:preliminary}
In this section, we will first provide a concise overview of pivoted QR factorization, Interpolative Decomposition (ID), and CUR decomposition. Then, we will introduce the CUR-ID algorithm, which forms the basis of our proposed method.

\textbf{Pivoted QR factorization.} For given a matrix $W\in\mathbb{R}^{m\times n}$ with real or complex entries, and set $m\geq n$ without loss of generality. The (compact) QR factorization then takes the form
\begin{equation}
    \mathop{W}\limits_{m\times n}\mathop{P}\limits_{n\times n}
    = \mathop{Q}\limits_{m\times n}\ \mathop{R}\limits_{n\times n},
\end{equation}
where $P$ is a permutation matrix, $Q$ is an orthogonal matrix and $R$ is upper triangular. The permutation matrix $P$ can be represented via a vector $J\subset [n]$ of indices that $P=I(:,J)$ where $I$ is the $n\times n$ identity matrix.

The QR-factorization is often computed via column pivoting, which results in factor $R$ satisfying various decay conditions\citep{golub2013matrix}, such as:
\begin{equation}
    \begin{aligned}
        R(k,k)^2\geq\sum\limits_{i=k}^{j}R(i, j)^2,
        \\
        j=k+1:n,k=1:n,
    \end{aligned}
\end{equation}

\textbf{Interpolative decomposition(ID) and CUR decomposition.} 
Generally, CUR decomposition approximates matrix $W$ by a product of three matrices $C$, $U$ and $R$, where matrices $C$ and $R$ consist of subset columns and rows from $W$ and $U$ is a small carefully constructed matrix to minimize the low-rank approximation error. Similary, ID approximates a matrix $W$ as a product of a matrix $C$ consisting a small subset of columns from $W$ and a coefficient matrix $X$.

From skeleton selection standpoint\citep{dong2023simpler}, given any arbitrary linearly independent column subset $C=W(:,J)(W\in \mathbb{R}^{m\times n}, J\subset [n])$, the rank-$\vert J\vert$ column ID of $W$ with respect to column skeletons can be formulated as
\begin{equation}
    \hat{W}_{*,J} \triangleq C(C^{\dag}W),
\end{equation}
where $CC^{\dag}$ is the orthogonal projector onto the spaning subspace of column skeletons. Analogously, the rank-$\vert K\vert$ column ID of $W$ with respect to row skeletons can be formulated as
\begin{equation}
    \hat{W}_{J,*} \triangleq (WR^{\dag})R,
\end{equation}
where row subset $R=W(K,:)(K\subset[m])$.

With both column and row skeletons, we can construct low-rank approximation in two-sided ID and CUR decomposition. We define $|K| = |J|$ and $S\triangleq W(K,J)$ be an invertible two-sided skeleton, such that two-sided ID
\begin{equation}
    \hat{W}_{I,J} \triangleq (CS^{-1})S(C^{\dag}W),
\end{equation}
and CUR decomposition
\begin{equation}
    \tilde{W}_{I,J} \triangleq C(C^\dag WR^\dag) R,
\end{equation}

Different from sampling-based methods that draw skeletons from proper probability distributions such as \citet{mahoney2009cur}, which utilized statistical leverage scores originating from statistics, \citet{voronin2017efficient} proposed a novel CUR-ID algorithm drawing skeleton selections via more deterministic pivoting. They demonstrated that a CUR decomposition could be constructed using a two-sided ID, which can itself be built from pivoted QR factorization. The matrices $C$ and $R$ are selected by two successive one-sided IDs. The idea behind this work is that the matrix $C$ can be directly obtained via ID, and then a subsequent full-rank ID on matrix $C$ yields an index vector needed to construct matrix $R$.

\textbf{Notation.} Given any matrix $W$ and (ordered) index set $J$ and $K$, $W(:, J)$ denotes the submatrix of $W$ consisting of columns from $W$ indexed by $J$. $W(K, J)$ denotes the submatrix obtained by rows and columns from $W$ indexed by $K$ and $J$ respectively\citep{golub2013matrix}.Given a positive integer $m$, the notation $[m]$ is defined as the set of the first $m$ natural numbers $\{1,2,\dots,m\}$. The notation $\dag$ denotes the Moore-Penrose pseudoinverse. In contrast to common conventions in computer science, all indices in this paper, unless otherwise specified, will begin from 1. This choice aligns with certain mathematical traditions and is made for consistency throughout the text.

\section{Methodology}
\subsection{Formulation of PMSS}
We present PMSS(\textbf{P}re-trained \textbf{M}atrices \textbf{S}keleton \textbf{S}election), a novel parameter efficient fine-tuning method designed to enhance the parameter efficiency of large language model while preserving linguistic and semantic information. We reparameterize the weight update matrix $\Delta W$ as the product of three matrices $C$,$U$ and $R$. Unlike LoRA, where both $A$ matrix initialized to zero and $B$ matrix initialized with Gaussian noise are trainable, our algorithm adopts a different way. Once the $C$ and $R$ matrices are initialized, they are frozen and no longer updated. Correspondingly, the $U$ matrix remains trainable throughout the training stage, which leads to computational efficiency while retaining the relevant structure from the pre-trained weights.
PMSS selects the most representative column and row skeletons of pre-trained matrices $W\in \mathbb{R}^{m\times n}$ to construct matrices $C\in \mathbb{R}^{m\times c}$ and $R\in\mathbb{R}^{r\times n}$. Consequently, we can represent $C$ and $R$ through compressed index vectors $K\subset [m]$ and $J\subset [n]$ respectively, further enhancing memory efficiency. The overview of PMSS is illustrated in Figure \ref{fig:overview}.

\textbf{Skeleton Selection.} We observe that different subspace initializations can impact the fine-tuning effectiveness of large language models. To capture the underlying structure (or skeleton) of the original weight more deterministicly, we employ a two-sided ID algorithm. Initially, we apply a one-sided ID to the original weight matrix $W$, identifying the column skeleton (i.e., matrix $C$) by performing a rank-c column-pivoted QR factorization. This yields the row index vector $J$, which is used to construct C.Subsequently, we perform successive one-sided ID on matrix $C$ through a full-rank column-pivoted QR factorization to derive the row skeleton, which forms the matrix $R$. Then, we can generate a row index vector $K$. Ultimately, we only need to explicitly retain the index vectors. The overall algorithm is summarized in Algorithm \ref{alg:PMSS algorithm}.

\textbf{Forward Pass.}
LoRA aims to reparameterize updates $\Delta W\in\mathbb{R}^{m\times n}$ of pre-trained matrix $W\in\mathbb{R}^{m\times n}$ in the form of low-rank approximation of adapter matrices $A\in \mathbb{R}^{r\times n}$ and $B\in \mathbb{R}^{m\times r}$ with rank $r\ll\min(m,n)$. LoRA's forward pass is:
\begin{equation}
    \label{eq:lora_forward}
    y=W'x=(W+BA)x,
\end{equation}
where $x\in \mathbb{R}^n$ is the input for the current layer, and $y\in \mathbb{R}^m$ is the output of the current layer and pre-activation input for the next layer. Both $A$ and $B$ matrices are trainable. 

Our proposed forward pass is:
\begin{equation}
    \begin{aligned}
        \label{eq:pmss_forward}
        y=W'x=(W+CUR)x, \\
        C=W(:,J),R=W(K,:),
    \end{aligned}
\end{equation}
where $K\in \mathbb{R}^{r}, K\subset [m]$, $J\in\mathbb{R}^c, J\subset [n]$. $K$ and $J$ are compressed index vectors, where each scalar value represents a selected row or column from the original pre-trained matrix. $C$ and $R$ are frozen after selection and $U$ retains trainable during training stage. We initialize the $U$ with zero to prevent any weight drift in the beginning.
We scale $\Delta Wx$ by $\frac{\alpha}{\max\{c,r\}}$, where $\alpha$ is a constant in $c,r$.

\begin{algorithm}[tb]
    \caption{PMSS Algorithm}
    \label{alg:PMSS algorithm}
    \textbf{Input}: Pretrained matrix $W\in \mathbb{R}^{m\times n}$. \\
    \textbf{Parameter}: Column number parameter $c$, row number parameter $r$, $min(m,n)\geq c \geq r$ WLOG.\\
    \textbf{Output}:  Column index set $J$ and row index set $K$. \\
    \begin{algorithmic}[1] 
    
    \STATE Perform a rank $c$ column pivoted QR factorization to get $WP := QR$ \\
    \STATE Define the column orderd index set $J$ via $I(:, J)=P$ \\
    \STATE Define an interpolation matrix $C := W(:, J(1:c))$
    \STATE Perform a full rank column pivoted QR factorization to get $CP^* := Q^*R^*$ \\
    \STATE Define the row orderd index set $K$ via $I(:, K) := P^*$ \\
    \STATE Partition $J:=J(1:c)$, $K := K(1:r)$ \\
    \STATE \textbf{Return} column index set $J$ and row index set $K$ \\

    \end{algorithmic}
    \end{algorithm}

\subsection{Fine-tuning Happens in Constraining Skeleton Subspaces}
In this subsection, we will compare the differences in gradient updates between LoRA and PMSS. Based on Equation \ref{eq:lora_forward}, during back-propagation, the gradient of weight matrix $W$ is:
\begin{equation}
    \nabla_{W'}\mathcal{L} = \frac{\partial \mathcal{L}}{\partial y}x^T,
\end{equation}
where $\mathcal{L}$ is the upstream loss and $\frac{\partial \mathcal{L}}{\partial y}$ denotes the partial derivative of $\mathcal{L}$ with respect to $y$.

In LoRA, the adapter matrices $A$ and $B$ are both trainable, and the gradients for these are computed separately as follows:
\begin{equation}
    \frac{\partial \mathcal{L}}{\partial A} = B^T\nabla_{W'}\mathcal{L},
    \frac{\partial \mathcal{L}}{\partial B} = \nabla_{W'}\mathcal{L}A^T,
\end{equation}
From the above equations, it is evident that the gradient computations for the matrices $A$ and $B$ in LoRA are mutually coupled and continuously evolving. However, in PMSS, the only trainable matrix is $U$, and its gradient can be computed as follows based on Equation \ref{eq:pmss_forward}:
\begin{equation}
    \frac{\partial \mathcal{L}}{\partial U} = C^T\nabla_{W'}\mathcal{L}R^T,
\end{equation}
We then find PMSS update matrix $U$ with SGD for every step $t$ by
\begin{equation}
    U_{t+1} \leftarrow U_{t} - \eta	C^T\nabla_{W'}\mathcal{L}_t R^T,
\end{equation}
where $\eta$ is the learning rate. Therefore, putting it to Equation \ref{eq:pmss_forward}, we reparameterize $\Delta W$ by
\begin{equation}
    \Delta W = -\eta CC^T(\sum\limits_{t=1}^{T}\nabla_{W'}\mathcal{L}_t)R^TR,
\end{equation}
Let $\sum\limits_{t=1}^{T}\nabla_{W'}\mathcal{L}_t$ be $M_T$ for convenience. This equation indicates the entire update is confined to the subspace spanned by $C$ and $R$. Although the matrices $C$ and $R$ are typically not orthogonal and thus $CC^T$ and $R^TR$ do not form strict projection matrices. However, $CC^T$ still constrains the rows of $M_T$ to the column space of $C$, and $R^TR$ constrains the columns of $M_T$ to the row space of $R$. Since $C$ and $R$ are selected from matrix $W$, the update is effectively confined to a constrained subspace spanned by the row and column skeletons of $W$. In contrast, FLoRA\citep{hao2024flora} demonstrates that the update $\Delta W$ in the vanilla initilization of LoRA can be approximated as: 
\begin{equation}
    \Delta W \approx -\eta (\sum\limits_{t=1}^{T}\nabla_{W'}\mathcal{L}_t)A_0^TA_0,
\end{equation}
where $A_0$ is the initialization of adapter matrix $A$ in LoRA. They reveal that LoRA updates can be viewed as performing random down and up projections to the gradient, whereas PMSS applies projections related to critical subspace of weight to the gradient.

\subsection{Parameter Efficiency and Low-Cost High-Rank Updates}

We demonstrate that our method achieves significant parameter efficiency compared to LoRA. For simplicity, let the number of layers for fine-tuning be $L_t$ and let the dimension of weights be $\mathbb{R}^{m\times n}$. For each layer, LoRA introduces a pair of trainable adapter matrices $A$ and $B$. The total number of trainable parameters $\Theta_{LoRA}$ in LoRA is determined by the rank $r_{LoRA}$ of adapter matrices:
\begin{equation}
    \Theta_{LoRA} = L_t\times(m+n)\times r_{LoRA},
\end{equation}

Similary, the total number of trainable parameters $\Theta_{PMSS}$ in PMSS is determined by $c_{PMSS}$ and $r_{PMSS}$ of adapter matrix $U$:
\begin{equation}
    \Theta_{PMSS} = L_t\times c_{PMSS}\times r_{PMSS},
\end{equation}
We observe PMSS's parameter efficiency arises from the fact that the number of trainable parameters is independent of the dimensions of the pre-trained weight matrix, $m$ and $n$, which are typically much larger in the large-scale model compared to the rank $r_{LoRA}$ in LoRA.

PMSS enables higher-rank updates than LoRA when the budget for trainable parameters is the same. Without loss of generality, we set parameter $c_{PMSS}$ equal to parameter $r_{PMSS}$ in PMSS. We further observe the rank of updates in PMSS 
\begin{equation}
    r_{PMSS}=\sqrt{(m+n)\times r_{LoRA}}\gg r_{LoRA},
\end{equation}
when $r_{LoRA} \ll \min(m,n)$. Even in resource-intensive environments, PMSS enables high-rank updates without increasing memory and computation costs compared with LoRA.

\subsection{Comparison with Other Works}
\label{sec:comparison}

Compared to works such as LoRA-XS or CURLoRA, which only focus on updates at low ranks and typically underperform LoRA, our algorithm emphasizes low-cost, high-rank updates instead.
Compared to selecting a set of abstract orthonormal bases via SVD, we argue that choosing elements directly from the original matrix provides greater interpretability.
In contrast to our method, CURLoRA focuses on mitigating catastrophic forgetting during the fine-tuning stage. They induce implicit regularization by probabilistically sampling inversely proportional to the row and column norms of the matrix, aiming to deviate from the original weight matrix. Their core idea is to induce implicit regularization by deviating as much as possible from the original pre-trained weight matrix. However, they overlook the self-expressive capability of the weights in large pre-trained language models, treating them merely as a means to enforce a certain form of regularization against LoRA. By selecting less significant features or even noise from the original matrix, their work risks capturing suboptimal or even deficient subspaces within the $C$ and $R$. This can lead to the loss of valuable information crucial for effective fine-tuning, thereby hindering the model's ability to adapt to new tasks efficiently.

\section{Experiments}
In this section, we conduct a series of experiments on various NLP benchmarks to showcase the efficiency of PMSS.
For convenience, we choose hyperparameter $c \text{ equal to } r$ in PMSS for all experimental settings.
All experiments are conducted on the NVIDIA H800(80G) GPUs. We will list implementation details and hyperparameters of these experiments in Appendix~\ref{sec:implementation} and \ref{sec:hyper}.

\subsection{DROP Benchmark}
\begin{table}
    \centering
    \resizebox{\columnwidth}{!}{

        \begin{tabular}{lccc}
            \hline
            \textbf{Model} & \textbf{Method}     & \textbf{\# Params\%} & \textbf{$F_1$ Score} \\
            \hline
            \multirow{14}{*}{LLaMA2$_{7B}$}
                           & FT$^\dag$           & 100\%                & 59.4                 \\
                           & Series$^\dag$       & 0.747\%              & 58.8                 \\
                           & Parallel$^\dag$     & 0.747\%              & 59.0                 \\
                           & LoRA$_{r=8}^\dag$   & 0.062\%              & 54.0                 \\
                           & LoRA$_{r=32}^\dag$  & 0.249\%              & 54.8                 \\
                           & LoRA$_{r=128}^\dag$ & 0.996\%              & 56.2                 \\
                           & CURLoRA$_{c,r=128}$ & 0.016\%             & 54.1                 \\
                           & CURLoRA$_{c,r=256}$ & 0.062\%             & 58.5                 \\
                           & CURLoRA$_{c,r=512}$ & 0.248\%              & 59.3                 \\
                           & CURLoRA$_{c,r=640}$ & 0.388\%              & 58.9                 \\
                           & PMSS$_{c,r=128}$    & 0.016\%             & 55.1                 \\
                           & PMSS$_{c,r=256}$    & 0.062\%             & \textbf{59.6}        \\
                           & PMSS$_{c,r=512}$    & 0.248\%              & \textbf{59.6}        \\
                           & PMSS$_{c,r=640}$    & 0.388\%              & 59.3        \\
            \hline
            \multirow{3}{*}{LLaMA2$_{13B}$}
                           & LoRA$_{r=8}^\dag$   & 0.050\%              & 61.0                 \\
                           & CURLoRA$_{c,r=128}$ & 0.010\%              & 64.3                 \\
                           & PMSS$_{c,r=128}$    & 0.010\%              & \textbf{66.9}        \\

            \hline
        \end{tabular}
    }

    \caption{Benchmark of various fine-tuning methods on the DROP dataset using LLaMA2 7B/13B models as the base model. We report $F_1$ score as metric and higher score is better. All results with $^\dag$ are taken from \citet{chen2024quanta}.
    }
    \label{tab:drop}
\end{table}

\begin{figure}[t]
    \includegraphics[width=\columnwidth]{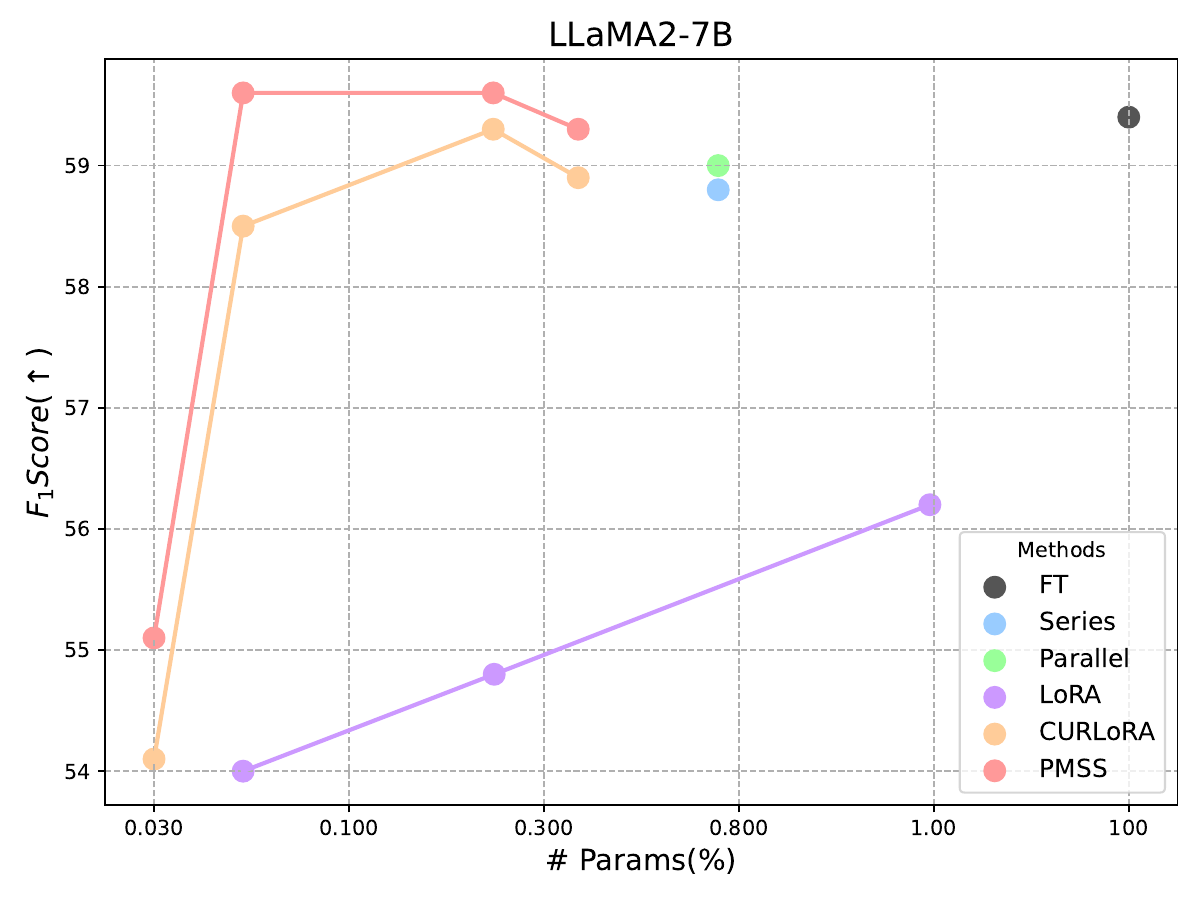}
    \caption{Benchmark of different fine-tuning methods on the DROP dataset. Illustration of the $F_1$ score (y-axis) with different numbers ratio(\%) of trainable parameters (x-axis) using LLaMA2-7B as the base model.
    }
    \label{fig:drop_fig}
\end{figure}

\begin{table*}
    \centering

    \resizebox{\textwidth}{!}{
        \begin{tabular}{lccccccccccc}
            \hline
            \textbf{Model} & \textbf{Method}    & \textbf{\#Params(\%)} & \textbf{BoolQ} & \textbf{PIQA}  & \textbf{SIQA}
                           & \textbf{HellaSwag} & \textbf{WinoGrande}   & \textbf{ARC-e} & \textbf{ARC-c} & \textbf{OBQA} & \textbf{AVG}                                                                                  \\
            \hline
            \multirow{2}{*}{LLaMA2$_{7B}$}
                           & LoRA$^\dag$        & 0.83                  & 69.8           & 79.9           & \textbf{79.5} & 83.6          & \textbf{82.6} & 79.8          & 64.7          & \textbf{81.0} & 77.6          \\
                           & CURLoRA            & 0.62                  & 69.0           & 80.1           & 78.4          & 81.5          & 77.5          & 78.4          & 63.5          & 75.8          & 75.5          \\
                           & PMSS               & 0.47                  & \textbf{70.8}  & \textbf{82.8}  & 78.2          & \textbf{88.4} & 80.9          & \textbf{82.5} & \textbf{68.0} & 80.2          & \textbf{79.0} \\

            \hline

            \multirow{3}{*}{LLaMA3$_{8B}$}
                           & LoRA$^\dag$        & 0.70                  & 70.8           & 85.2           & \textbf{79.9} & 91.7          & 84.3          & 84.2          & 71.2          & 79.0          & 80.8          \\
                           & CURLoRA            & 0.13                  & 73.5           & 87.5           & 79.0          & 94.4          & \textbf{83.4} & \textbf{90.2} & 78.5          & 84.6          & 83.9          \\
                           & PMSS               & 0.13                  & \textbf{73.8}  & \textbf{87.8}  & 78.9          & \textbf{94.6} & 84.2          & 89.6          & \textbf{78.7} & \textbf{85.8} & \textbf{84.2} \\
            \hline
        \end{tabular}
    }
    \caption{Comparison of LLaMA-2 7B and LLaMA-3 8B with various PEFT methods on eight commonsense reasoning datasets.
        All results with $^\dag$ of baseline methods are taken from \citet{liu2024dora}.
        We report accuracy as metric and higher score is better.
    }
    \label{tab:commonsense_result}
\end{table*}

\textbf{Datasets and Models.} We first conduct experiments on the DROP dataset\citep{dua2019drop}, a challenging English reading comprehension benchmark that requires models to perform discrete reasoning over paragraphs. Through experiments using LoRA with varying ranks and subspace similarity analysis, \citet{chen2024quanta} demonstrate that DROP is a representative example of a higher intrinsic rank dataset compared to other NLP datasets, such as RTE dataset\citep{wang2019superglue}. We evaluate PMSS against LoRA and several baseline methods which include full fine-tuning, Series adapter (Series)\citep{houlsby2019parameter}, Parallel adapter (Parallel)\citep{he2022towards} and CURLoRA\citep{fawi2024curlora} by fine-tuning LLaMA-2 7B/13B\citep{touvron2023llama}.

\textbf{Results.} As shown in Table \ref{tab:drop}, PMSS consistently outperforms than other fine-tuning methods. We observe that PMSS achieves performance comparable to and in some cases exceeding, full fine-tuning while training only a small fraction of the parameters. This demonstrates the effectiveness of our method's high-rank updates. To investigate the performance of these methods with the ranks scaling, we conducted experiments across varying ranks of PMSS and CURLoRA on LLaMA2-7B. As illustrated in Figure \ref{fig:drop_fig}, we present a line graph depicting the variation of $F_1$ score with respect to ranks (i.e., the number of trainable parameters) for LoRA, CURLoRA, and PMSS. As shown in the figure, the performance of PMSS, CURLoRA, and LoRA improve as the ranks increase, underscoring both the necessity and effectiveness of high-rank updates. As the rank increases, PMSS exhibits rapid performance gains, achieving superior results compared to other methods, including LoRA and CURLoRA, with minimal trainable parameters (0.062\%). Conversely, LoRA only achieves subpar performance, though it improves with increased parameters.

\subsection{Commonsense Reasoning}
\label{sec:commonsense_experi}

\textbf{Datasets and Models.} We fine-tune our models on COMMONSENSE170K\citep{hu2023llm}, a comprehensive dataset of various commonsense reasoning questions. Eight different commonsense reasoning datasets are used for evaluation, including BoolQ\citep{clark2019boolq}, PIQA\citep{bisk2020piqa}, SIQA\cite{sap2019socialiqa}, HellaSwag\citep{zellers2019hellaswag}, WinoGrande\citep{sakaguchi2021winogrande}, ARC-e, ARC-c\citep{clark2018think} and OBQA\citep{ling2017obqa}. All commonsense reasoning tasks are formulated as multiple-choice or Yes/No questions, where the models are required to select the most appropriate answers and rationales. We evaluate PMSS against LoRA and CURLoRA by fine-tuning LLaMA-2 7B\citep{touvron2023llama} and LLaMA-3 8B\citep{dubey2024llama}. 

\begin{table}
    \centering
    \resizebox{\columnwidth}{!}{

        \begin{tabular}{lccc}
            \hline
            \textbf{Model} & \textbf{q$_{proj}$}     & \textbf{k$_{proj}$} & \textbf{v$_{proj}$} \\
            \hline
        LLaMA2-7B & $4096\times 4096$ & $4096\times 4096$ & $4096\times 4096$ \\
        LLaMA3-8B & $4096\times 4096$ & $4096\times \textbf{1024}$ & $4096\times \textbf{1024}$ \\

            \hline
        \end{tabular}
    }
    \caption{Comparison of module dimensions on LLaMA2-7B and LLaMA3-8B.}

    \label{tab:module_dimension}
\end{table}

\textbf{Results.} 
The main results are reported in Table \ref{tab:commonsense_result}.
PMSS outperforms LoRA and CURLoRA on most metrics for the LLaMA2-7B and LLaMA3-8B models.
On LLaMA2-7B, PMSS outperforms LoRA and CURLoRA by 1.4\% and 3.5\% average accuracy scores. On LLaMA3-8B, PMSS exceeds LoRA and CURLoRA by 3.4\% and 0.3\% average accuracy scores. 

However, PMSS underperforms LoRA on 3 out of 8 evaluation metrics on LLaMA2-7B, and underperforms LoRA on 2 metrics on LLaMA3-8B. It may be due to the inherently lower rank of commonsense reasoning tasks compared to more complex tasks(e.g., math). As a result, LoRA remains competitive in such tasks.
Additionally, on LLaMA3-8B, our method performs closely to CURLoRA. As shown in Table \ref{tab:module_dimension}, the asymmetric structure of weight($k_{proj} \text{ and } v_{proj}$) in LLaMA3-8B may make it easier to capture the critical subspace. We explore this point further in the ablation studies.

\subsection{Math Reasoning}
\label{sec:math_experi}
\begin{table}
    \centering
    \resizebox{\columnwidth}{!}{
        \begin{tabular}{lcccc}
            \hline
            \textbf{Model} & \textbf{Method} & \textbf{Params} & \textbf{GSM8K} & \textbf{MATH}  \\
            \hline
            \multirow{5}{*}{LLaMA2$_{7B}$}
                           & Full FT$^\dag$  & 6738M           & 49.05          & 7.22           \\
                           & LoRA$^\dag$     & 320M            & 42.30          & 5.50           \\
                           & PiSSA$^\dag$    & 320M            & 53.07          & 7.44           \\
                           & CURLoRA         & 56M             & 54.51          & 9.30           \\
                           & PMSS            & 56M             & \textbf{55.19} & \textbf{9.74}  \\

            \hline
            \multirow{5}{*}{Mistral$_{7B}$}
                           & Full FT$^\dag$  & 7242M           & 67.02          & 18.6           \\
                           & LoRA$^\dag$     & 168M            & 67.70          & 19.68          \\
                           & PiSSA$^\dag$    & 168M            & 72.86          & \textbf{21.54} \\
                           & CURLoRA         & 87.5M           & 72.40          & 20.40          \\
                           & PMSS            & 87.5M           & \textbf{73.31} & 21.34          \\
            \hline
            \multirow{5}{*}{Gemma$_{7B}$}
                           & Full FT$^\dag$  & 8538M           & 71.34          & 22.74          \\
                           & LoRA$^\dag$     & 200M            & 74.90          & 31.28          \\
                           & PiSSA$^\dag$    & 200M            & 77.94          & \textbf{31.94} \\
                           & CURLoRA         & 49M             & 76.65          & 30.20          \\
                           & PMSS            & 49M             & \textbf{78.01} & 30.60          \\

            \hline
        \end{tabular}
    }
    \caption{
    Math reasoning evaluation results for LLaMA2-7B, Mistral-7B and Gemma-7B on math reasoning benchmarks. All results with $^\dag$ of baseline methods are taken from \citet{meng2024pissa}. We report accuracy as metric and higher score is better.}
    \label{tab:math}
\end{table}

\textbf{Datasets and Models.} We train our models on MetaMathQA dataset\citep{yu2023metamath}, which comprises 395K samples augmented from other math instruction tuning datasets such as GSM8K\citep{cobbe2021training} and MATH\citep{hendrycks2021measuring}, with higher diversity and complexity. We select GSM8K and MATH as the test datasets. We select LLaMA-2 7B\citep{touvron2023llama}, Mistral-7B-v0.1\citep{jiang2023mistral} and Gemma-7B\citep{team2024gemma} as base models. We evaluate PMSS against full fine-tuning, LoRA, PiSSA, and CURLoRA as baseline methods.

\textbf{Results.}
Table \ref{tab:math} presents the evaluation results on the GSM8K and MATH benchmarks. The results show that PMSS outperforms LoRA and CURLoRA and even surpasses full fine-tuning with a small fraction of parameters. On LLaMA2-7B, PMSS outperforms PiSSA by +2.12/2.30 on GSM8k and MATH. However, On both the Mistral-7B and Gemma-7B, PMSS outperforms PiSSA on the GSM8K but falls short of PiSSA on the MATH. This suggests that PiSSA's core idea of utilizing the principal singular components of the weight matrices is effective. Nevertheless, PMSS still performs comparably to PiSSA with much fewer parameters(about 18\%-52\%), demonstrating the potential scalability of PMSS when handling even complex tasks like math reasoning.

\subsection{Ablation Study}
\begin{table}
    \centering
        \begin{tabular}{ccc}
            \hline
            \textbf{Method} & \textbf{\#Params(\%)} & \textbf{AVG}  \\
            \hline
            Random         & 0.13                  & 83.9          \\
            \hline
            CURLoRA         & 0.0082                & 78.6          \\
            Random         & 0.0082                & 78.9          \\
            PMSS            & 0.0082                & \textbf{79.5} \\

            \hline
        \end{tabular}
    \caption{Ablation study results on commonsense reasoning using LLaMA3-8B. We also report the result of random selection when the rank is kept the same as in Section~\ref{sec:commonsense_experi}.}
    \label{tab:ablation_commonsense}
\end{table}

\begin{table}
    \centering
        \begin{tabular}{lcccc}
            \hline
            \textbf{Method} & \textbf{Params} & \textbf{GSM8K} & \textbf{MATH} \\
            \hline
            Random         & 56M             &        53.68        &      9.38         \\
            \hline
            CURLoRA         & 3.5M                &      46.70          &       6.76        \\
            Random         &      3.5M           &    45.64            &       7.04        \\
            PMSS            &       3.5M        &  \textbf{47.69} & \textbf{7.38} \\

            \hline
        \end{tabular}
    \caption{Ablation study results on math reasoning using LLaMA2-7B. We also report the result of random selection when the rank is kept the same as in Section~\ref{sec:math_experi}.}
    \label{tab:ablation_math}
\end{table}

In this subsection, we provide ablation study results to empirically demonstrate how different skeleton selection strategies impact the experimental outcomes. In previous experiments, we set a relatively high rank (e.g., 512). High-rank updates may provide an inherent advantage by enabling the model to learn more effectively. When selecting a sufficient number of rows and columns from the original weight matrix to construct the skeletons, even a random selection method can yield good results, as it effectively covers the row and column spaces' information of the weight matrix. 
We use random selection and CURLoRA as a baseline. We adhere to the hyperparameters used in previous experiments, modifying only the learning rate. We also report the results of the random selection method at higher-rank updates, using the exact same hyperparameter settings as in the previous experiments.

\textbf{Ablation Study on Commonsense Reasoning.} As shown in Table \ref{tab:ablation_commonsense}, we present the average scores of the eight commonsense evaluation sets. PMSS consistently outperforms baseline methods at lower ranks. We also report the result of random selection at the same high rank and find it performs closely to CURLoRA and PMSS, which may be attributed to the asymmetric structure of weight matrices on LLaMA3-8B,

\textbf{Ablation Study on Math Reasoning.} As shown in Table \ref{tab:ablation_math}, we present the evaluation result on GSM8K and MATH sets. PMSS consistently outperforms baseline methods across all rank settings.

\section{Conclusion}
In this paper, we introduce PMSS, a novel fine-tuning method designed to enhance parameter efficiency while preserving the semantic and linguistic information of weight. Our method effectively overcomes the low-rank limitation of LoRA and enables high-rank updates at a low cost. Experimental results demonstrate that PMSS outperforms LoRA and other fine-tuning methods with much less trainable parameters. PMSS is expected to demonstrate superior learning capacity, especially in handling complex tasks.

\newpage
\section{Limitation}
We have leveraged the inherent world knowledge embedded in the model’s weights. However, to better adapt the model to specific downstream tasks, task-specific knowledge should also be incorporated, which remains for future research.
Due to time and objective conditions, it is still unclear whether PMSS is effective for other specific tasks, such as logical reasoning tasks. 
Additionally, the effectiveness of this method on models with larger parameter scales (e.g. 70B) remains to be verified.

\bibliography{coling_latex}

\clearpage
\appendix

\onecolumn

\section{Implementation Details}
\label{sec:implementation}

\subsection{DROP dataset}
\textbf{Implementation Details.} We conduct experiments across different ranks and compare PMSS's performance with LoRA and CURLoRA, highlighting the impact of high-rank updates. We follow the setting of \citet{chen2024quanta}, selecting 2,000 samples from the training set of the DROP dataset as our training set, 800 samples as the validation set, and 1,200 samples from the validation set of DROP as the test set. $F_1$-score is used as the evaluation metric for measuring the closeness of the model's output with the ground truth. The best checkpoint from the validation set is loaded as the final model for testing. \citet{chen2024quanta} select the number of epoch parameters arbitrarily between 3 and 6. For a fair comparison, we standardize our experiments to run for 3 epochs.

\subsection{Commonsense Reasoning Dataset}

\textbf{Implementation Details.}
We first fine-tune the model on the joint COMMONSENSE170K dataset and evaluate the fine-tuned model on eight downstream commonsense tasks. 
We follow the setting of \citet{liu2024dora}, splitting the COMMONSENSE170K dataset in a train set of 170,020 samples and a validation set of 400 samples. We optimize the hyperparameters on the validation set and load the best checkpoint on the validation set as the final model for evaluation. Accuracy is reported as a metric.
We use the implementation of LLM-Adapters\citep{hu2023llm} and standardize our experiments to run for 3 epochs.

\subsection{Math Reasoning Dataset}
\textbf{Implementation Details.} In math reasoning experiments, we use the implementation of PiSSA\citep{meng2024pissa} and follow the setting of them while only fine-tuning the learning rate. All models are first trained on a subset containing 100K data points from the MetaMathQA. We load the last checkpoint as the final model for evaluation. Then the fine-tuned models are evaluated on GSM8K and MATH datasets to assess their capabilities in solving mathematical problems as specific downstream tasks. Accuracy is reported on the GSM8K and MATH datasets. All models are fine-tuned for only one epoch.

\section{Case Studies}
\label{sec:case}
In this section, we present examples of three different tasks to help readers gain a deeper understanding of the specifics of the Three different benchmarks. Additionally, the process output of the Math task(GSM8K) further highlights the validity of the theoretical assumptions discussed in the main text.
\subsection{DROP}
\begin{mdframed}[linewidth=1pt,linecolor=black]
\textbf{[An example in DROP]} 

\textbf{Context:} \newline
\hspace{3cm}As of the census of 2000, there were 325,957 people, 149,937 households, and 94,460 families residing in the county.  The population density was 570 people per square mile (220/km2).  There were 182,467 housing units at an average density of 319 per square mile (123/km2).  The racial makeup of the county was 92.65\% Race (United States Census), 4.18\% Race (United States Census) or Race (United States Census), 0.22\% Race (United States Census), 0.77\% Race (United States Census), 0.03\% Race (United States Census), 1.14\% from Race (United States Census), and 1.02\% from two or more races.  4.34\% of the population were Race (United States Census) or Race (United States Census) of any race. 89.7\% spoke only English language at home; 4.4\% spoke the Spanish language, 1.3\% German language, and 1.0\% French language at home.

\textbf{Question:} Which group is smaller for the county according to the census: people or families?

\textbf{Answeanswerilies.} 

\end{mdframed}
\begin{mdframed}[linewidth=1pt,linecolor=black]
\textbf{[PMSS Reasoning by LlaMA2-7B]}


\textbf{[Prediction: families.]} (\textcolor{green}{Right})
\end{mdframed}

\begin{mdframed}[linewidth=1pt,linecolor=black]
\textbf{[CURLoRA Reasoning by LlaMA2-7B]}

\textbf{[Prediction: people.]} (\textcolor{red}{Wrong})
\end{mdframed}

\subsection{ARC-e}
\begin{mdframed}[linewidth=1pt,linecolor=black]
\textbf{[An example in ARC-e]} 

\textbf{Context:} \newline
\hspace{3cm}Below is an instruction that describes a task. Write a response that appropriately completes the request. Instruction: Please choose the correct answeanswerhe question: 
Copper (Cu) reacts with oxygen (O) to form copper oxide (CuO). The properties of CuO are most likely

\textbf{Options:} \newline
\textbf{Answer1:} different from copper or oxygen. \newline
\textbf{Answer2:} similar to both copper and oxygen.   \newline
\textbf{Answer3:} similar only to copper.  \newline
\textbf{Answer4:} similar only to oxygen. \newline
\textbf{Question:} Which one is the correct answeanswerswer1/answer2/answer3/answer4)

\textbf{Answer: the correct answeanswernswer1.} 

\end{mdframed}
\begin{mdframed}[linewidth=1pt,linecolor=black]
\textbf{[PMSS Reasoning by LlaMA3-8B]}

\textbf{[Prediction: the correct answer is answer1.]} (\textcolor{green}{Right})
\end{mdframed}

\begin{mdframed}[linewidth=1pt,linecolor=black]
\textbf{[CURLoRA Reasoning by LlaMA3-8B]}

\textbf{[Prediction: the correct answer is answer2.]} (\textcolor{red}{Wrong})
\end{mdframed}

\subsection{GSM8K}
\begin{mdframed}[linewidth=1pt,linecolor=black]
\textbf{[An example in GSM8K]} 

\textbf{Context:} \newline
\hspace{3cm}question: Below is an instruction that describes a task. Write a response that appropriately completes the request.
Helga was the fastest clog dancer in all of Slovenia. With both hands at her sides, she could tap her right foot at a rate of 300 taps per minute while simultaneously tapping her left foot at a rate of 250 taps per minute.  When she raised her arms, her tap rate slowed down to 200 taps per minute with each foot.

\textbf{Question:} If she dances a total of 5 minutes, with her arms raised during only 2 of those minutes, what would be the combined total number of times that she taps both of her feet?

\textbf{Answer: 2450.} 

\end{mdframed}
\begin{mdframed}[linewidth=1pt,linecolor=black]
\textbf{[PMSS Reasoning by Gemma-7B]}

\textbf{Reasoning Process:} 
\begin{itemize}
    \item In the first 3 minutes, Helga taps her right foot at a rate of 300 taps per minute, so she taps her right foot 300 * 3 = 900 times.

    \item In the first 3 minutes, Helga taps her left foot at a rate of 250 taps per minute, so she taps her left foot 250 * 3 = 750 times.

    \item In the first 3 minutes, Helga taps both feet a total of 900 + 750 = 1650 times.

    \item In the next 2 minutes, Helga taps her right foot at a rate of 200 taps per minute, so she taps her right foot 200 * 2 = 400 times.

    \item In the next 2 minutes, Helga taps her left foot at a rate of 200 taps per minute, so she taps her left foot 200 * 2 = 400 times.

    \item In the next 2 minutes, Helga taps both feet a total of 400 + 400 = 800 times.

    \item In the total 5 minutes, Helga taps both feet a total of 1650 + 800 = 2450 times.

    The answeanswer2450
\end{itemize}

\textbf{[Prediction: 2450.]} (\textcolor{green}{Right})
\end{mdframed}

\begin{mdframed}[linewidth=1pt,linecolor=black]
\textbf{[CURLoRA Reasoning by Gemma-7B]}

\textbf{Reasoning Process:} 
\begin{itemize}
    \item In the first 3 minutes, Helga dances with her arms at her sides, so she taps each foot at a rate of 300 taps per minute and 250 taps per minute, respectively.\

    \item In the first 3 minutes, she taps her right foot 3 minutes x 300 taps per minute = 900 times.

    \item In the first 3 minutes, she taps her left foot 3 minutes x 250 taps per minute = 750 times.

    \item In the first 3 minutes, she taps both feet a total of 900 taps + 750 taps = 1650 times.

    \item In the next 2 minutes, Helga dances with her arms raised, so she taps each foot at a rate of 200 taps per minute..

    \item In the next 2 minutes, she taps her right foot 2 minutes x 200 taps per minute = 400 times.

    \item In the next 2 minutes, she taps her left foot 2 minutes x 200 taps per minute = 400 times.

    \item In the next 2 minutes, she taps both feet a total of 400 taps + 400 taps = 800 times.

    \item In the last minute, Helga dances with her arms at her sides again, so she taps each foot at a rate of 300 taps per minute and 250 taps per minute, respectively.

    \item In the last minute, she taps her right foot 1 minute x 300 taps per minute = 300 times.

    \item In the last minute, she taps her left foot 1 minute x 250 taps per minute = 250 times.

    \item In the last minute, she taps both feet a total of 300 taps + 250 taps = 550 times.

    \item In the last minute, she taps both feet a total of 550 times.
\end{itemize}

\textbf{[Prediction: 550.]} (\textcolor{red}{Wrong})
\end{mdframed}

\section{Hyperparamaters}
\label{sec:hyper}

\begin{table}
    \centering

        \begin{tabular}{lcc}
            \hline
             & Hyperparameters & PMSS\quad CURLoRA         \\

            \hline
            &Batch Size      & 4                      \\
            &Optimizer       & AdamW                  \\
            &Scheduler       & Linear Scheduler       \\
            &Weight Decay    & 0                      \\
            &Dropout         & 0                      \\
            &Modules         & q\_proj, v\_proj       \\
            &Number of GPUs  & 2                      \\
            &Epochs          & 3                     \\
            \hline

            Model &$c,r,\alpha$ & Learning Rate \\
            \hline
            \multirow{4}{*}{LLaMA2-7B} &128-128-128  &      3e-4    \\
            &256-256-256  &     4e-4     \\
            &512-512-512  &      7e-5    \\
            &640-640-640  &        7e-5   \\

            \hline
            LLaMA2-13B &128-128-128  &      1e-3    \\
            \hline

        \end{tabular}

    \caption{Hyperparameters used for DROP dataset for PMSS and CURLoRA on LLaMA2-7B and LLaMA2-13B. }
    \label{tab:hyper_drop}
\end{table}

\begin{table}
    \centering
        \begin{tabular}{lcc}
            \hline
             Hyperparameters & LLaMA2-7B &  LLaMA2-13B         \\
            \hline
            Batch Size      & \multicolumn{2}{c}{16}                      \\
            Warmup Steps   & \multicolumn{2}{c}{100} \\
            Optimizer       & \multicolumn{2}{c}{AdamW}                  \\
            Scheduler       & \multicolumn{2}{c}{Linear Scheduler}       \\
            Weight Decay    & \multicolumn{2}{c}{0}                      \\
            Dropout         & \multicolumn{2}{c}{0.05}                      \\
            Modules         & \multicolumn{2}{c}{q\_proj, v\_proj,k\_proj, up\_proj, down\_proj}       \\
            Number of GPUs  & \multicolumn{2}{c}{1}                      \\
            Epochs          & \multicolumn{2}{c}{3}                     \\
            \hline

            $c,r,\alpha$(PMSS)    & 448-448-896 & 256-256-512 \\
            Learning Rate(PMSS) & 1e-4 & 1e-4 \\
            \hline

            $c,r,\alpha$(CURLoRA)    & 512-512-1024 & 256-256-512 \\
            Learning Rate(CURLoRA) & 2e-4 & 2e-4 \\
            \hline

        \end{tabular}

    \caption{Hyperparameter configurations of PMSS and CURLoRA for  LLaMA2-7B, and LLaMA3-8B on the commonsense reasoning
tasks.}
    \label{tab:hyper_commonsense}
\end{table}

\begin{table}
    \centering

        \begin{tabular}{lccc}
            \hline
             Hyperparameters & PMSS &  Random & CURLoRA         \\
            \hline

            $c,r,\alpha$    & \multicolumn{3}{c}{64-64-128}\\

            Learning Rate  & \multicolumn{3}{c}{7e-4} \\
            \hline

        \end{tabular}

    \caption{Ablation study on the commonsense reasoning tasks. We report hyperparameter configurations of PMSS, Random and CURLoRA for LLaMA3-8B.}
    \label{tab:hyper_ablation_commonsense}
\end{table}

\begin{table}
    \centering

        \begin{tabular}{lccc}
            \hline
             Hyperparameters & LLaMA2-7B &  Mistral-7B   & Gemma-7B        \\
            \hline
            Batch Size      & \multicolumn{3}{c}{128}                      \\
            Warmup Ratio   & \multicolumn{3}{c}{0.03} \\
            Optimizer       & \multicolumn{3}{c}{AdamW}                  \\
            Scheduler       & \multicolumn{3}{c}{Cosine}       \\
            Weight Decay    & \multicolumn{3}{c}{0}                      \\
            Dropout         & \multicolumn{3}{c}{0}                      \\
            Modules         & \multicolumn{3}{c}{q\_proj, v\_proj,k\_proj, o\_proj, gate\_proj, up\_proj, down\_proj}       \\
            Number of GPUs  & \multicolumn{3}{c}{1}                      \\
            Epochs          & \multicolumn{3}{c}{3}                     \\

            $c,r,\alpha$    & 512-512-1024 & 640-640-1280 & 512-512-1024 \\
            Learning Rate(PMSS) & 2e-4 & 1e-3 & 7e-4  \\

            Learning Rate(CURLoRA) & 4e-4 & 7e-4  & 7e-4 \\
            \hline

        \end{tabular}

    \caption{Hyperparameter configurations of PMSS and CURLoRA for LLaMA2-7B, Mistral-7B and Gemma-7B on the math reasoning tasks.}
    \label{tab:hyper_math}
\end{table}

\begin{table}
    \centering

        \begin{tabular}{lccc}
            \hline
             Hyperparameters & PMSS &  Random & CURLoRA         \\
            \hline

            $c,r,\alpha$    & \multicolumn{3}{c}{128-128-256}\\

            Learning Rate  & 2e-3 & 1e-3 & 2e-3\\
            \hline

        \end{tabular}

    \caption{Ablation study on the commonsense reasoning tasks. We report hyperparameter configurations of PMSS, Random and CURLoRA for LLaMA2-7B.}
    \label{tab:hyper_ablation_commonsense}
\end{table}

\end{document}